\pdfoutput=1

\documentclass[11pt]{article}

\usepackage[]{acl}

\usepackage{times}
\usepackage{latexsym}

\usepackage[inline]{enumitem}

\usepackage{amsfonts}

\usepackage{amsmath}

\usepackage{bm}

\usepackage{booktabs}

\usepackage{multirow}

\usepackage{graphicx}
\usepackage{subcaption}

\usepackage{lipsum}
\usepackage{tabulary}
\usepackage{geometry}
\usepackage{blindtext}
\usepackage{graphicx}
\usepackage{caption,tabularx,booktabs}

\usepackage{caption}
\usepackage{subcaption}

\usepackage[T1]{fontenc}

\usepackage[utf8]{inputenc}

\usepackage{microtype}

\title{Multi Task Learning For Zero Shot Performance Prediction of Multilingual Models}

\author{Kabir Ahuja\textsuperscript{1} \thanks{\hspace{0.1cm} Equal contribution}  \quad Shanu Kumar\textsuperscript{2} \footnotemark[1] \quad Sandipan Dandapat\textsuperscript{2} \quad Monojit Choudhury\textsuperscript{1}\\
\textsuperscript{1} Microsoft Research, India \\
\textsuperscript{2} Microsoft R\&D, Hyderabad, India \\
{\tt \small \{t-kabirahuja,shankum,sadandap,monojitc\}@microsoft.com}
}

\begin{document}
\maketitle
\begin{abstract}
Massively Multilingual Transformer based Language Models have been observed to be surprisingly effective on zero-shot transfer across languages, though the performance varies from language to language depending on the pivot language(s) used for fine-tuning. In this work, we build upon some of the existing techniques for predicting the zero-shot performance on a task, by modeling it as a multi-task learning problem. We jointly train predictive models for different tasks which helps us build more accurate predictors for tasks where we have test data in very few languages to measure the actual performance of the model. Our approach also lends us the ability to perform a much more robust feature selection, and identify a common set of features that influence zero-shot performance across a variety of tasks.

\end{abstract}

\section{Introduction}

Multilingual models like mBERT \cite{devlin-etal-2019-bert} and XLM-R \cite{conneau-etal-2020-unsupervised} have been recently shown to be surprisingly effective for zero-shot transfer \cite{pires-etal-2019-multilingual} \cite{wu-dredze-2019-beto}, where on fine-tuning for a task on one or a few languages, called {\em pivots}, they can perform well on languages unseen during training. The zero-shot performance however, is often not uniform across the languages and the multilingual models turn out to be much less effective for low resource languages \cite{wu-dredze-2020-languages, lauscher-etal-2020-zero} and the languages that are typologically distant from the pivots \cite{lauscher-etal-2020-zero}. What affects the zero-shot transfer across different languages is a subject of considerable interest and importance~\cite{wang2019cross, pires-etal-2019-multilingual, wu-dredze-2019-beto, lauscher-etal-2020-zero}, however there is little conclusive evidence and a few papers even show contradictory findings.

\citet{lauscher-etal-2020-zero} recently, showed that it is possible to predict the zero shot performance of mBERT and XLM-R on different languages by formulating it as a regression problem, with pretraining data size and typological similarities between the pivot and target languages as the input features, and the performance on downstream task as the prediction target. Along similar lines \citet{srinivasan2021predicting} and \citet{dolicki2021analysing} explore zero-shot performance prediction with a larger set of features and different regression techniques.

However, the efficacy of these solutions are severely limited by the lack of training data, that is, the number of languages for which performance metrics are available for a given task. For instance, for most tasks in the popular  XTREME-R \cite{ruder2021xtreme} benchmark, there are data points for 7-11 languages. This not only makes zero-shot performance prediction a challenging problem, but also a very important one because for practical deployment of such multilingual models, one would ideally like to know its performance for all the languages the model is supposed to handle. As \citet{srinivasan2021predicting} shows, accurate performance predictors can also help us build better and fairer multilingual models by suggesting data labeling strategies.

In this work, we propose multi-task learning~\cite{zhang2017multitask} as an approach to mitigate training-data constraints and consequent over-fitting of the performance predictors to tasks and/or datasets. 
The contributions of our work are fourfold.
 \textit{First,} we experiment with different multi-task learning approaches, such as Group Lasso \cite{yuan2006model}, Collective Matrix Factorization \cite{cortes2018cold}, Multi-Task Deep Gaussian Process Regression \cite{NIPS2007_66368270} and Meta Agnostic Meta Learning \cite{finn2017model} for 11 tasks. We observe an overall 10\% reduction in performance prediction errors compared to the best performing single-task models. The gains are even stronger when we just consider the tasks with very few data points ($\leq 10$), where we see a 20\% drop in the mean absolute errors.
  \textit{Second,} an interesting consequence of modelling this problem via multi-task learning is that we are able to predict performance on low resource languages much more accurately, where in some cases single-task approaches may perform even worse than the simple averaging baselines.
 \textit{Third,} apart from the features used for zero-shot performance prediction in the previous work \cite{lauscher-etal-2020-zero, srinivasan2021predicting, dolicki2021analysing}, we also utilize metrics quantifying the quality of multilingual tokenizers as proposed in \cite{rust-etal-2021-good} as features in our predictive models, which turn out to have strong predictive power for certain tasks. To the best of our knowledge, our work is the first to explore the impact of tokenizer quality specifically on zero-shot transfer. 
 And \textit{fourth,} our multi-task framework in general lends us with a much more robust selection of features affecting the zero-shot performance. This, in turn, lets us investigate the critical open question on what influences the zero-shot performances across languages more rigorously. As we shall see, our findings corroborate some of the previous conclusions, while others are extended or annulled.

\section{Background and Related Work}
\label{sec:rel_work}


\noindent 
\textbf{Zero Shot Transfer.} Multilingual models like mBERT~\cite{devlin-etal-2019-bert} and XLM-R \cite{conneau-etal-2020-unsupervised} have shown surprising effectiveness in zero-shot transfer, where fine-tuning the MMLM on a task in some source language often leads to impressive performance on the same task in other languages as well without explicitly training on them. \citet{pires-etal-2019-multilingual} first observed this phenomenon for NER \cite{tjong-kim-sang-2002-introduction, tjong-kim-sang-de-meulder-2003-introduction, levow-2006-third} and POS tagging \cite{nivre2018universal} tasks. Concurrently, \citet{wu-dredze-2019-beto} also showed this surprisingly cross lingual transfer ability of mBERT additionally on tasks like Document Classification \cite{SCHWENK18.658},  Natural Language Inference \cite{Conneau2018xnli} and Dependency Parsing \cite{nivre2018universal}.

\noindent
\textbf{Factors Affecting Zero Shot Transfer.} \citet{pires-etal-2019-multilingual} showed that vocabulary memorization played little role in zero-shot generalization as language pairs with little word piece overlap also exhibited impressive crosslingual performance. \citeauthor{wang2019cross} arrived at a similar conclusion by training BERT on an artificially generated language to zero out the word overlap with the target languages, and observed only minor drops in the performance compared to training the model on English. On the contrary \citet{wu-dredze-2019-beto}, observed strong correlations between the sub-word overlap and the zero-shot performance in four out of five tasks. 

 \citet{wu-dredze-2020-languages} showed that mBERT performed much worse for zero-shot transfer to low resource languages (i.e., less pre-training data) than high resource ones on POS Tagging, NER and Dependency Parsing tasks. \citet{lauscher-etal-2020-zero} also had a similar observation on tasks like XNLI and XQuAD \cite{artetxe2020cross}, though they found that the zero-shot performance on NER, POS tagging and Dependency Parsing tasks might not strictly depend on the pre-training size and could be better explained by different linguistic relatedness features like syntactic and phonological similarities between the language pair. Similar dependence on the typological relatedness such as word order had also been observed by \citet{pires-etal-2019-multilingual}.

\noindent
\textbf{Performance Prediction.} 
Prior work has explored predicting the performance of machine learning models from unlabelled data by either measuring (dis)agreements between multiple classifiers \cite{platanios2014estimating, platanios2017estimating} or by utilizing underlying information about data distribution \cite{domhan2015speeding}. In the context of NLP \citet{birch-etal-2008-predicting} explored predicting the performance of a Machine Translation system by utilizing different explanatory variables for the language pairs. \citet{lin-etal-2019-choosing} proposed a learning to rank approach to choose transfer languages for cross lingual learning using several linguistic and dataset specific features. 

Recently, there has been an interest in predicting the performance of NLP models without actually training or testing them, by formulating it as a regression problem. \citet{xia-etal-2020-predicting} showed that using experimental settings for an NLP experiment as inputs it is possible to accurately predict the performance on different languages and model architectures.\citet{ye-etal-2021-towards} extended this work by proposing methods to do a fine-grained estimation of the performance as well as predicting well-callibrated confidence intervals.
Specifically predicting the zero-shot performance of MMLMs was first explored in \citet{lauscher-etal-2020-zero}, where they used a linear regression model to estimate the cross-lingual transfer performance based on pre-training data size and linguistic relatedness features. \citet{srinivasan2021predicting} tackled this problem by utilizing XGBoost Regressor for the prediction along with a larger set of features. \citet{dolicki2021analysing} explored individual syntactic features for zero-shot performance prediction instead of working with aggregate similarity values, and showed about 2 to 4 times gain in performance. We extend all of these works by considering a multi-task learning approach, where performance prediction in a task utilizes not only the data available for that task, but also the patterns observed for other tasks. 

\section{Problem Setup}

We begin by defining the multi-task performance prediction problem and then describe the different linguistic and MMLM specific features used. 

\subsection{Multi-Task Performance Prediction Problem}

Consider a pre-trained multilingual model $\mathcal{M}$, trained using self supervision on a set of languages $\mathcal{L}$. Let $\mathfrak{T}$ be the set of downstream NLP tasks, $\mathcal{P}$ be the set of pivot (source) languages for which training data is available for the downstream tasks for fine-tuning and $\mathcal{T}$ be the set of target languages for which validation/test data is available. Note that $\mathcal{P} \subset \mathcal{L}$ and $\mathcal{T} \subseteq \mathcal{L}$. We use the zero-shot setting similar to \citet{lauscher-etal-2020-zero} which enforces $\mathcal{P}$ and $\mathcal{T}$ to be disjoint sets\footnote{Though beyond the scope of the current work, it is possible to extend this to a few-shot setting as discussed in~\citet{srinivasan2021predicting}.}, i.e., $\mathcal{P} \cap \mathcal{T} = \emptyset$.

We then define $y_{p,t}^{\mathcal{M}, \mathfrak{t}} \in \mathbb{R}$ as the zero-shot performance on language $t \in \mathcal{T}$ on finetuning $\mathcal{M}$ on task $\mathfrak{t} \in \mathfrak{T}$ in pivot language $p \in \mathcal{P}$. 
Let $x_{p,t}^{\mathcal{M}} \in \mathbb{R}^{n}$ be the $n$-dimensional feature vector representing the corresponding train-test configuration. 
Since for our experiments we train and evaluate the performance prediction for a single model at a time, we will simplify the notations to $y_{p,t}^{\mathfrak{t}}$ and $x_{p,t}$.

The predictor model can then be defined as the function $f_{\Theta, \Phi} : \mathbb{R}^n \times \mathfrak{T} \rightarrow \mathbb{R}$, where $\Theta \in \mathbb{R}^{d_{g}}$ denotes the shared parameters across the tasks and the task specific parameters are given by $\Phi \in \mathbb{R}^{d_{s} \times |\mathfrak{T}|}$. The objective function for training such a predictor model can be defined as:
\begin{equation}
    \begin{split}
        J(\Theta, \Phi) &= \sum_{\mathfrak{t} \in \mathfrak{T}} \sum_{p \in \mathcal{P}} \sum_{t \in \mathcal{T}} \| f(x_{p,t}, \mathfrak{t}; \Theta, \Phi) - y_{p,t}^{\mathfrak{t}} \|_2^2 \\
        &+ \lambda_{g} \|\Theta\|_{1} + \lambda_{s} \|\Phi\|_{1,1} + \lambda_{group} \|\Phi\|_{1,q}\\
    \end{split}
    \label{eq:mt_obj}
\end{equation}

The second and third terms regularize the global and task specific parameters independently, while the last term, $l_1/l_q$ norm with $q > 1$, ensures a block sparse selection of the task specific parameters. This term ensures a multi-task learning behavior even when there are no parameters shared across the tasks (i.e., $\Theta = \emptyset$) through selection of common features across the tasks. Setting $\Theta = \emptyset$ and $\lambda_{group} = 0$ leads to the single task setup of \citet{lauscher-etal-2020-zero} and \citet{srinivasan2021predicting}.

\subsection{Features}
\label{sec:feats}

We divide the set of features into two higher level categories, viz. the pairwise features defined for the pivot and target that measure the typological relatedness of the languages, and the individual features defined for the target language reflecting the state of its representation in $\mathcal{M}$. 

\subsubsection{Pairwise Features}
Instead of directly using the different typological properties of the the two languages as features, we use the pairwise relatedness to avoid feature explosion.


\noindent \textbf{Subword Overlap} : We define the subword overlap as the percentage of unique tokens that are common to the vocabularies of both the pivot and target languages. Let $V_p$ and $V_t$ be the subword vocabularies of $p$ and $t$. The subword overlap is then defined as :

\begin{equation}
    o_{sw}(p, t) = \frac{|V_p \cap V_t|}{|V_p    \cup V_t|}
\end{equation}


\noindent \textbf{Similarity between Lang2Vec vectors}: 
Following \citet{lin-etal-2019-choosing} and \citet{lauscher-etal-2020-zero}, we compute the typological relatedness between $p$ and $t$ from the linguistic features provided by the URIEL project \cite{littell-etal-2017-uriel}. 
We use syntactic ($s_{syn}(p,t)$), phonological similarity ($s_{pho}(p,t)$), genetic similarity ($s_{gen}(p,t)$) and geographic distance ($d_{geo}(p,t)$). For details, please see \citet{littell-etal-2017-uriel}.

\subsubsection{Individual Features}



\noindent \textbf{Pre-training Size}: 
We use the $log_{10}$ of the size (in words) of the pre-training corpus in the target language, $\text{SIZE}(t)$, as a feature. 

\noindent \textbf{Rare Typological Traits}: \citet{srinivasan2021predicting} proposed this metric to capture the rarity of the typological features of a language in the representation of $\mathcal{M}$. Every typological feature in WALS database is ranked based on the amount of pre-training data for the languages that contain the feature. For the language $t$, Mean Reciprocal Rank (MRR) of all of its features is then calculated and used as a feature -- $\text{WMRR}(t)$.

\noindent \textbf{Tokenizer Features} : In their recent work, \citet{rust-etal-2021-good} proposed two metrics, viz. tokenizer's \textit{fertility} and proportion of continued words, to evaluate the quality of multilingual tokenizers on a given language. For target $t$, they define the tokenizer's fertility, $\text{FERT}(t)$, as the average number of sub-words produced for every tokenized word in $t$'s corpus. On the other hand, the proportion of continued words, $\text{PCW}(t)$, measures how often the tokenizer chooses to continue a word across at least two tokens. They show that the multilingual models perform much worse on a task than their monolingual counterparts when the values of these metrics are higher for the multilingual tokenizer. We include $\text{FERT}(t)$ and $\text{PCW}(t)$ as features.\\

\noindent An important thing to note here is that the we do not use identity of a language as a feature while training the models, hence the performance prediction models are capable of generating predictions on new languages unseen during training. However, if the features of the new languages deviate significantly from the features seen during training, the predictions are expected to be less accurate as also observed in \citet{xia-etal-2020-predicting, srinivasan2021predicting} and is one of the main reasons for exploring a multi-task approach.



\section{Approaches}

We extensively experiment with a wide-array of multi-task as well as single-task regression models to provide a fair comparison between different approaches to zero-shot performance prediction.

\subsection{Baselines}

\noindent \textbf{Average Score Within a Task (AWT)} : The performance for a pivot-target pair ($p$ , $t$) on a task $\mathfrak{t}$ is approximated by taking the average of the performance on all other target languages (pivot being fixed) in the same task $\mathfrak{t}$, i.e., $f(x_{p,t}, \mathfrak{t}) = \frac{1}{|\mathcal{T}| - 1}\sum_{t'\in \mathcal{T} - \{t\}}y_{p,t'}^{\mathfrak{t}}$.

\noindent \textbf{Average Score across the Tasks (AAT)} : Here instead of averaging over all the target languages within a task, we approximate the performance on a given target language by averaging the scores for that language across the other tasks, i.e., $f(x_{p,t}, \mathfrak{t}) = \frac{1}{|\mathfrak{T}| - 1}\sum_{\mathfrak{t}' \in \mathfrak{T} - \{\mathfrak{t}\}}y_{p,t}^{\mathfrak{t'}}$.

\subsection{Single Task Models}

\noindent \textbf{Lasso Regression}: \citet{lauscher-etal-2020-zero} train different linear regression models for each task. Along similar lines, we experiment with linear regression, but also add an L1 regularization term, as we observed it usually leads to better predictors.

\noindent \textbf{XGBoost Regressor}: As shown in \citet{srinivasan2021predicting}, XGBoost \cite{xgboost} generally obtains impressive performance on this task, and hence we include it in our experiments as well. 

\subsection{Multi Task Models}

\noindent \textbf{Group Lasso}: $l_1/l_q$ norm based block-regularization has been shown to be effective for multi-task learning in the setting of multi-linear regression \cite{yuan2006model,argyriou2008convex}. For each task, consider separate linear regression models represented by the weight matrix $\Phi \in \mathbb{R}^{n \times |\mathfrak{T}|}$. The $l_1/l_q$ regularization term is given as: $\|\Phi\|_{1,q} = \sum_{j = 1}^{n}(\sum_{\mathfrak{t} = 1}^{|\mathfrak{T}|}{|\Phi_{j\mathfrak{t}}|^{q}})^{1/q}$ , where $\Phi_{j\mathfrak{t}}$ denotes the weight for the feature $j$ in the task $\mathfrak{t}$. For $q > 1$, minimizing this term pushes the $l_q$-norms corresponding to the weights of a given feature across the tasks to be sparse, which encourages multiple predictors to share similar sparsity patterns. In other words, a common set of features is selected for all the tasks. We use $q = 2$ for the group regularization term.

Since this can be restrictive in certain scenarios, some natural extensions to Group Lasso, such as Dirty Models \cite{dirty_model} and Multi Level Lasso \cite{multi_level_lasso}, have been proposed that separate out the task specific and global parameters. We experimented with these methods and observed equivalent or worse performance compared to Group Lasso. 

\noindent \textbf{Collective Matrix Factorization (CMF) with Side Information}: 
Low rank approximation for the task weights matrices forms one family of methods for multi-task learning \cite{zhang2017multitask,pong2010trace,ando2005framework}.
As a direct analogue with collaborative filtering, here we can think of the tasks as \textit{users} and pivot-target pairs as \textit{items}. Consider the matrix $\mathbf{Y} \in \mathbb{R}^{|\mathfrak{T}| \times |\mathcal{P} \times \mathcal{T}|}$, where each element of the matrix correspond to $y_{p,t}^{\mathfrak{t}}$. We can then decompose the matrix into task and language-pair specific factors as
\begin{equation}
    \mathbf{Y} \sim \mathbf{T}\mathbf{L}^{T}
\end{equation}
where $\mathbf{T} \in \mathbb{R}^{|\mathfrak{T}| \times d_{latent}}$ and $\mathbf{L} \in \mathbb{R}^{|\mathcal{P} \times \mathcal{T}| \times d_{latent}}$ are the task and language-pair factor matrices, and $d_{latent}$ is the number of factors.

Additionally, in order to incorporate the feature information about the language pairs as discussed in section \ref{sec:feats}, we incorporate Collective Matrix Factorization approach \cite{cortes2018cold}. It incorporates the attribute information about items and/or users in the factorization algorithm by decomposing the language-pair feature matrix $\mathbf{X} \in \mathbb{R}^{|\mathcal{P} \times \mathcal{T}| \times n}$ as
   $\mathbf{L}\mathbf{F}^T$,
such that $\mathbf{L}$ is shared across both decompositions. This helps to learn the latent representations for the pivot-language pairs from the task-wise performance as well as different linguistic and MMLM specific features\footnote{Note that we can use a similar approach for providing side information for the tasks as well.}. In relation to Equation \ref{eq:mt_obj}, we can think of task factors $\mathbf{T}$ to correspond to the task specific parameters $\Phi$, language-pair factors $\mathbf{L}$ as the shared parameters $\Theta$ and the predictor model as  $f(x_{p,t}, \mathfrak{t}; \Theta, \Phi) = (\mathbf{T}\mathbf{L}^T)_{(p,t), \mathfrak{t}}$. Both $\mathbf{L}$ and $\mathbf{T}$ are regularized seperately, but there is no group regularization term ($\lambda_{group} = 0$).

\citet{ye-etal-2021-towards} also uses a Tensor Factorization approach for performance prediction which is similar to our CMF method. However, they train separate models for each task and factorize over metric specific attributes instead for a fine-grained prediction.

\noindent \textbf{Multi-Task Deep Gaussian Process Regression (MDGPR)}: We use the multi-task variant of Gaussian Processes proposed in ~\citet{NIPS2007_66368270} and utilize deep neural networks to define the kernel functions as in Deep GPs \cite{pmlr-v51-wilson16}. For comparison, we also report the scores of the single-task variant of this method which we denote as DGPR. See Appendix (section \ref{sec:gpr_maml}) for details.

Apart from these we also explore other multi-task methods like Model Agnostic Meta Learning (MAML) \cite{finn2017model}, details of which we leave in the appendix (section \ref{sec:gpr_maml}).

\section{Experimental Setup}
In this section, we discuss our test conditions, datasets  and training parameters for the different experiments.

\subsection{Test Conditions}

We consider two different test conditions: Leave One Language Out (LOLO) and Leave Low Resource Languages Out (LLRO).

\noindent
\textbf{Leave One Language Out}:
LOLO is a popular setup for multilingual performance prediction \cite{lauscher-etal-2020-zero, srinivasan2021predicting}, where for a given task, we choose a target language and move all of its instances from the prediction dataset to the test data. The models are then trained on the remaining languages and evaluated on the unseen test language. This is done for all the target languages available for a task, and the Mean Absolute Error (MAE) across languages is reported. In the multi-task setting we evaluate on one task at a time while considering the rest as helper tasks for which the entire data is used including the test language\footnote{Note that this is a reasonable relaxation to make as it is closer to the real world use case where we would have the evaluation data for some languages in the standard tasks and would like to utilize that to make predictions on the same languages for the new ftask.}.

\noindent
\textbf{Leave Low Resource Languages Out}:
Through this evaluation strategy we try to emulate the real world use case where we only have test data available in high resource languages such as English, German and Chinese, and would like to estimate the performance on under-represented languages such as Swahili and Bengali. We use the language taxonomy provided by \citet{joshi-etal-2020-state} to categorize the languages into six classes (0 = low to 5 = high) based on the number of resources available. We then move languages belonging to class 3 or below to our test set and train the models on class 4 and 5 languages only. Similar to LOLO, here too we allow the helper tasks to retain all the languages.

\begin{table*}[]
\scalebox{0.6}{
\begin{tabular}{@{}cccccccccccc@{}}
\toprule
\textbf{MMLM} & \textbf{Task}    & \textbf{$|\mathcal{T}|$} & \multicolumn{2}{c}{\textbf{Baselines}}                       & \multicolumn{3}{c}{\textbf{Single Task Models}} & \multicolumn{4}{c}{\textbf{Multi Task Models}}                           \\\cmidrule(lr){4-5}\cmidrule(lr){6-8}\cmidrule(lr){9-12}
                 &&                                     & \textbf{Average within Task} & \textbf{Average across Tasks} & \textbf{Lasso}         & \textbf{XGBoost} & \textbf{DGPR}      & \textbf{Group Lasso} & \textbf{CMF}  & \textbf{MDGPR} & \textbf{MAML}   \\\midrule
\multirow{14}{4em}{XLMR}& \textbf{MLQA}    & 7                                   & 2.92                         & 2.26                          & 4.33                   & 2.91 & 3.26                   & \textbf{2.21}                 & 2.66 & 2.96            & 4.89            \\
&\textbf{PAWS}    & 7                                   & 3.34                         & 0.9                        & \textbf{0.8}                    & 1.28    & 1.27               & 1.32                 & 1.39          & 2.71            & 6.62            \\
&\textbf{XCOPA}   & 8                                   & 4.52                         & 5.91                          & 2.42                   & 4.16             & 4.73      & 2.69                 & 2.03          & \textbf{1.96}   & 6.28            \\
&\textbf{TyDiQA}  & 9                                   & 4.29                       & 5.48                        & 5.89                 & 5.63                & 6.56 & 5.04               & 5.88        & \textbf{4.61} & 4.96          \\
&\textbf{XQUAD}   & 10                                  & 4.90                       & 4.22                        & 4.54                 & 6.56             & 4.13    & 4.16               & 3.86        & \textbf{3.15} & 6.85          \\
&\textbf{LAReQA}  & 10                                  & 2.10                       & \textbf{1.51}               & 1.53        & 1.56          & 1.78       & 1.52               & 1.87        & 2.69          & 8.22          \\
&\textbf{MewsliX} & 10                                  & 16.61                      & 15.48                       & 15.70                & 21.16        & 15.66        & 13.73              & 14.62       & 10.07         & \textbf{9.33} \\
&\textbf{XNLI}    & 14                                  & 3.07                       & 2.07                        & 1.97        & \textbf{1.53}            & 2.16     & 2.17               & 2.17        & 3.54          & 4.55          \\
&\textbf{WikiANN} & 32                                  & 15.22                      & 11.61                       & 10.14                & 10.26           & 12.64     & 10.92              & 11.36       & \textbf{9.15} & 13.19         \\
&\textbf{Tatoeba} & 35                                  & 8.69                       & 8.68                       & \textbf{5.82}                 & 7.14              & 6.80   & 5.83      & 6.08        & 8.09          & 9.72          \\
&\textbf{UDPOS}   & 48                                  & 10.15                      & 7.65                       & 7.52                 & \textbf{5.12}    & 6.02    & 7.72               & 7.89        & 5.88          & 10.71\\
\cmidrule(lr){2-12}
&\textbf{Average} & 19 & 6.89 & 5.98 & 5.51 & 6.12 & 5.91 & 5.21 & 5.44 & \textbf{4.98} & 7.76\\
&\textbf{Average ($|\mathcal{T}| \leq 10$)} & 9 & 5.53 & 5.11 & 5.03 & 6.18 & 5.34 & 4.38 & 4.62 & \textbf{4.02} & 6.73\\
\midrule
\midrule
\multirow{2}{4em}{mBERT}&\textbf{Average} & 19 & 8.69 & 6.57 & 5.55 & 6.86 & 6.10 & 5.45 & \textbf{5.08} & 5.12 & 8.14\\
&\textbf{Average ($|\mathcal{T}| \leq 10$)} & 9 & 6.96 & 5.64 & 4.99 & 6.54 & 5.73 & 4.44 & \textbf{4.18} & 4.53 & 7.51\\
\bottomrule
\end{tabular}
}
\caption{\label{tab:results_xlmr} Mean Absolute Error (scaled by 100 for readability) for LOLO for different approaches across tasks. We also report the average MAE across all tasks (``Average'') and for tasks which has less than or equal to 10 languages (``Average ($|\mathcal{T}| \leq 10$)'').  Task-wise results for mBERT can be found in the Appendix (table \ref{tab:results_mbert})}
\end{table*}

\subsection{Tasks and Datasets}
We use the following 11 tasks provided in XTREME \cite{hu2020xtreme} and XTREME-R \cite{ruder2021xtreme} benchmarks: \begin{enumerate*}
    \item \textbf{Classification}: XNLI \cite{Conneau2018xnli} , PAWS-X \cite{Yang2019paws-x}, and XCOPA \cite{ponti-etal-2020-xcopa}
    \item \textbf{Structure Prediction}: UDPOS \cite{nivre2018universal}, and NER \cite{Pan2017}
    \item \textbf{Question Answering}: XQUAD \cite{artetxe2020cross}, MLQA \cite{Lewis2020mlqa}, and TyDiQA-GoldP \cite{Clark2020tydiqa}
    \item \textbf{Retrieval}: Tatoeba \cite{Artetxe2019massively}, Mewsli-X \cite{botha-etal-2020-entity, ruder2021xtreme}, and LAReQA \cite{roy-etal-2020-lareqa}
\end{enumerate*}

All of these datasets have training data present only in English i.e. $\mathcal{P} = \{\texttt{en}\}$, and majority of the tasks have fewer than 10 target languages. 



\subsection{Training Details}
We train and evaluate our performance prediction models for \textbf{mBERT} (\textit{bert-base-multilingual-cased}) and \textbf{XLM-R} (\textit{xlm-roberta-large}). For training XGBoost, we used 100 estimators with a maximum depth of 10. For Group Lasso, we used the implementation provided in the MuTaR software package\footnote{\url{https://github.com/hichamjanati/mutar}}, and used a regularization strength of $0.01$. We optimized CMF's objective function using Alternating Least Squares (ALS), used 5 latent factors with a regularization parameter equal to $0.1$, and used the Collective Matrix Factorization python library\footnote{\url{https://github.com/david-cortes/cmfrec}}. In case of MDGPR, we used Radial Basis Function as the kernel and a two-layer MLP for learning latent features, with 50 and 10 units followed by ReLU activation. We set the learning rate and epochs as 0.01 and 200, and implemented it using GPyTorch\footnote{\url{https://gpytorch.ai/}}.


\section{Results and Discussion}

\begin{figure}[b]
    \includegraphics[width=0.45\textwidth]{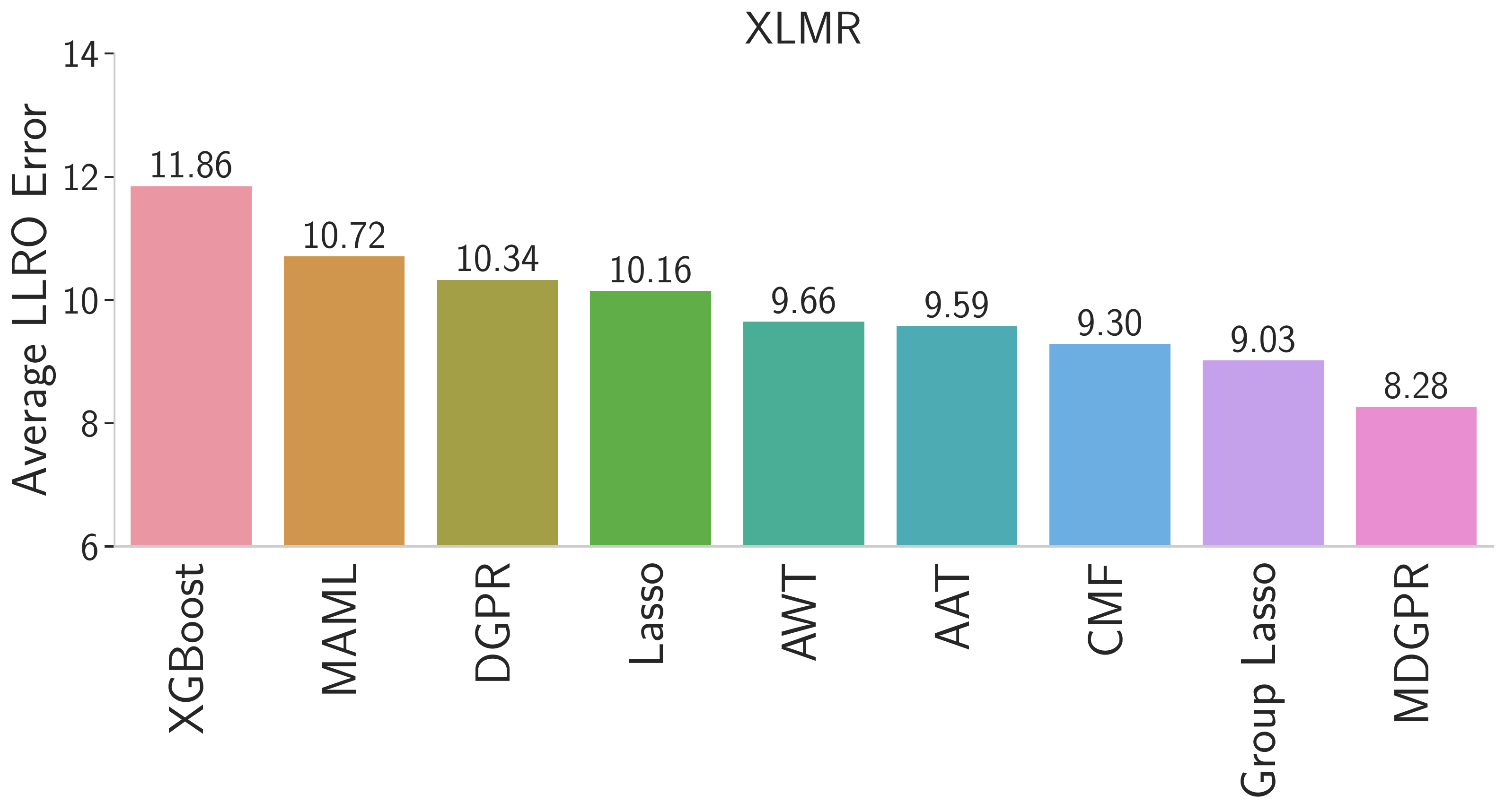}
    \caption{Leave Low Resource Out (LLRO) results for XLMR}
    \label{fig:xlmr_lowres}
\end{figure}

\subsection{LOLO Results}

Table \ref{tab:results_xlmr} shows MAE (in \%) for LOLO for different single-task and multi-task models on the tasks. For XLMR, we observe that multi-task models, primarily MDGPR, often outperform the best single-task models by significant margins, and for tasks like MewsliX we even see about 36\% reduction in MAE. Overall, we see about 10\% drop in LOLO errors on average for MDGPR compared to the best performing single-task model i.e. Lasso Regression. As expected, the benefit of multi-task learning is even more prominent when we consider the tasks for which only a few ($\leq 10$) data points are available. Here we see about 20\% reduction in errors. For mBERT as well, we have similar observations, except that CMF performs slightly better than MDGPR.

Note that the \textit{Average across task} baseline is quite competitive and performs better than single-task XGBoost and MAML in average, and better than all models for LAReQA.

\begin{figure}
    \centering
    \includegraphics[width=0.49\textwidth]{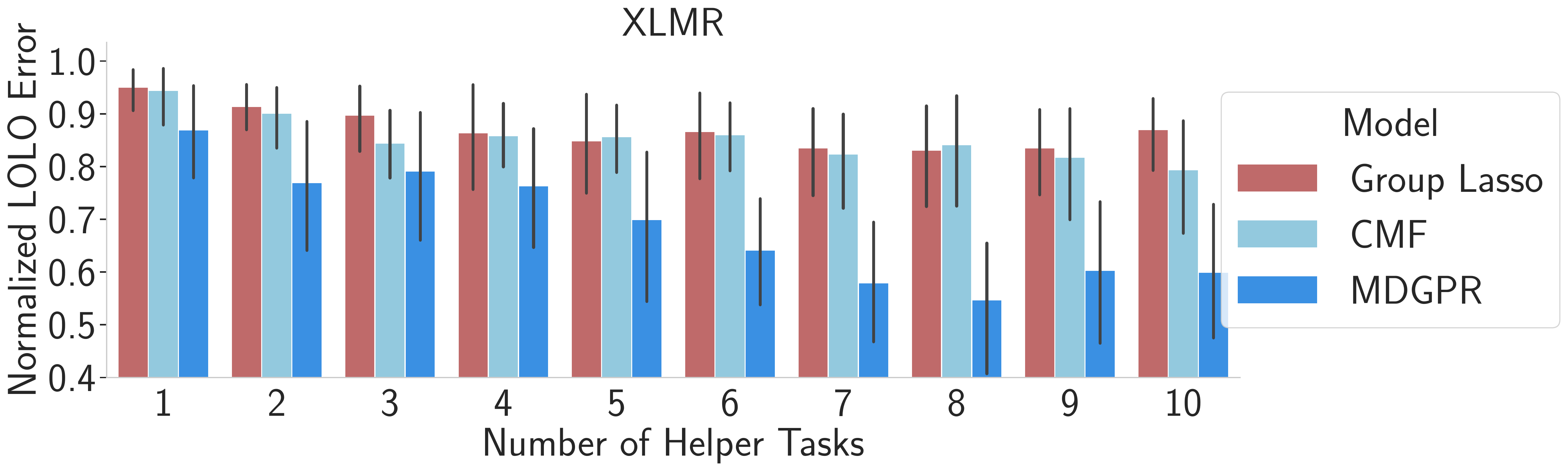}
    \caption{Number of helper tasks vs. LOLO MAE. Errors for different model types (Group Lasso, CMF and MDGPR) and tasks are scaled by diving them by the maximum error value.}
    \label{fig:xlmr_num_helpers}
\end{figure}

Figure \ref{fig:xlmr_num_helpers} plots the dependence of the number of helper tasks on the performance of the multi-task models. As expected, MAE decreases as helper tasks increase, especially for MDGPR and CMF. 
On a related note, the Pearson Correlation coefficient between MAE and number of tasks a target language is part of is found to be $-0.39$, though the trend in this case is not as clear.

\subsection{LLRO Results}
Predicting the performance on low resource languages, for which often standard training and test datasets are not available, can be an important use case where multi-task performance prediction can be helpful. Figure \ref{fig:class_dist} in appendix shows the class-wise \cite{joshi-etal-2020-state} distribution of languages for the tasks that we consider in our experiments. As one would expect, for most tasks, test data is available for languages belonging to class-4 and class-5. Training performance prediction models without any task to transfer from can therefore, possibly lead to poor generalization on the low resource languages. On the other hand, for the same reason - lack of test data, building accurate predictors for low-resource languages is necessary.

MAE values for the LLRO evaluation setup are shown in figure \ref{fig:xlmr_lowres} for XLMR. Results for mBERT follow similar trends and are reported in the Appendix (figure \ref{fig:mbert_lowres}). For both XLMR and mBERT we observe that the three main multi-task models -- Group Lasso, CMF and MDGPR -- outperform the single-task models and baselines. Interestingly, for XLMR, the single task models XGBoost and Lasso perform even worse than the \textit{Average within Tasks} baseline. Overall we see around 18\%  and 11\% drop in MAE for Group Lasso over the best performing single-task model, for XLMR and mBERT respectively. 

\subsection{Feature Importance}
An interesting consequence of zero-shot performance prediction is that the models can be directly used to infer the correlation (and possibly causation) between linguistic relatedness and pre-training conditions and zero-shot transferability. Multi-task learning, in this context, help us make more robust inferences, as the models are less prone to overfitting to a particular task or dataset. 

\begin{figure}
    \centering
    \includegraphics[width=0.49\textwidth]{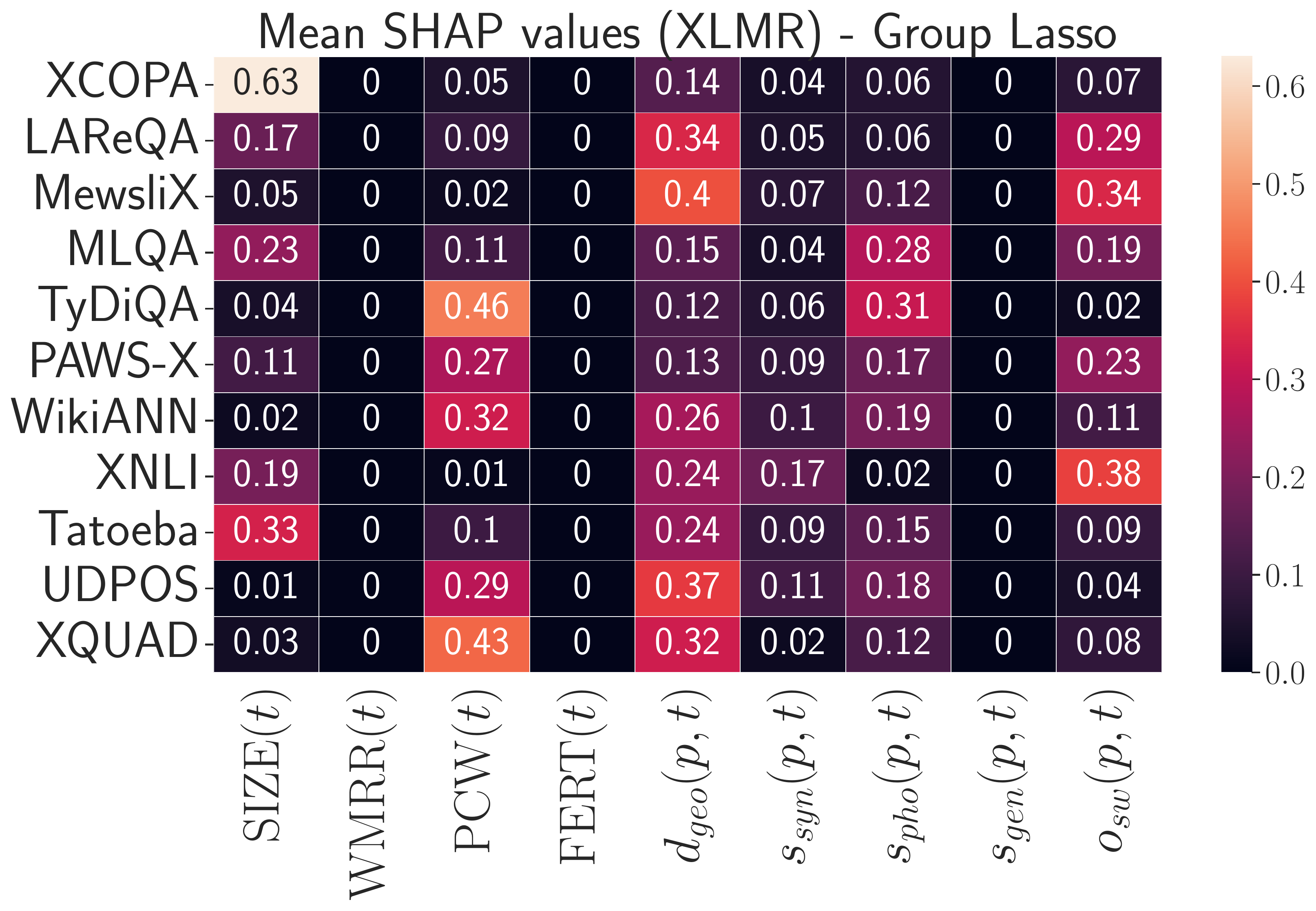}
    \caption{Task-wise mean SHAP values of different features for the Group Lasso model trained on XLMR zero-shot performance data. Higher value implies stronger effect.}
    \label{fig:xlmr_shap}
\end{figure}

Figure \ref{fig:xlmr_shap} shows the SHAP values of the  features for the Group Lasso model trained on XLMR's zero-shot performance data. As expected for Group Lasso, we see a block-sparsity behavior among the tasks. Features such as Rare Typological Traits ($\text{WMRR(t)}$), Tokenizer's Fertility ($\text{FERT}(t)$) and Genetic Similarity ($s_{gen}(p,t)$) are ignored in all the tasks.  In contrast, for the single-task lasso regression (Figure \ref{fig:lasso_xlmr_shap} in Appendix), we see different sets of features selected for different tasks, which for the scale at which we operate, might not be indicative of the actual factors that affect the zero-shot performance in these tasks.

\noindent
\textbf{Subword Overlap.} Among the features that get selected for all tasks, we observe that Subword Overlap ($o_{sw}(p,t)$) typically gets higher importance in retrieval (LAReQA and MewsliX) and sentence classification tasks (PAWS-X, XNLI). Since the retrieval tasks that we consider, as described in \citet{ruder2021xtreme}, measure the alignment between the cross lingual representations of semantically similar sentences, having a shared vocabulary between the languages can leak information from one to another \cite{wu-dredze-2019-beto} which might improve the retrieval performance. Interestingly, if we compare this with the feature importance scores for the single task lasso model (Figure \ref{fig:lasso_xlmr_shap} in Appendix), we do see MewsliX task getting higher importance for the subword overlap, but LAReQA gets virtually zero SHAP value for this feature, showcasing how single-task models can misinterpret two similar tasks as requiring very different features. Our observation reinforce the generally held notion that vocabulary overlap between the pivot and target is beneficial for zero-shot transfer \cite{wu-dredze-2019-beto}, especially for retrieval tasks, though some studies have argued otherwise~\cite{pires-etal-2019-multilingual, wang2019cross}. 

\noindent
\textbf{Tokenizer Features.} For structure prediction (UDPOS and WikiAnn) and question answering (XQUAD and TyDiQA) tasks that require making predictions for each token in the input, we see that the tokenizer feature, $\text{PCW}(t)$, receive a higher SHAP value. In contrast, for single-task lasso, here too we do not observe high importance of this feature across these related tasks. \citet{rust-etal-2021-good} note that languages such as Arabic where mBERT's multilingual tokenizer was found to be much worse than it's monolingual counterpart, there was a sharper drop in performance of mBERT compared to the monolingual model for QA, UDPOS and NER tasks than for sentiment classification. We believe that XLMR's surprisingly worse performance than mBERT for Chinese and Japanese on UDPOS might be correlated with it's significantly worse tokenizer for these languages based on the \textit{fertility} ($\text{FERT}$) and \textit{Percentage Continued Words} ($\text{PCW}$) feature values (see Appendix \ref{sec:mmlm_tokenizer} for exact values). The high SHAP values for $\text{PCW}(t)$ further strengthen our belief\footnote{Note that \citet{rust-etal-2021-good} shows the importance of tokenizer metrics for the case where the multilingual models are fine-tuned on the target language, whereas we analyze their importance for zero-shot transfer.}.

\noindent
\textbf{Pre-training Size.} Similar to the findings of \citet{lauscher-etal-2020-zero}, we observe that pre-training corpus size has low SHAP value, and therefore, lower importance for lower level tasks such as UDPOS and NER, and higher SHAP values for higher level tasks like XNLI. Additionally, we extend their observations to tasks such as XCOPA, Tatoeba, MLQA and LAReQA where  pre-training size seem to play a significant role in the performance prediction. Again, compared to single Lasso Regression model, we see a different selection pattern: Pre-training size receives a high SHAP value for UDPOS while for XNLI it is negligible. This neither fully conforms with our observations on the multi-task feature selections, nor with the previous work \cite{lauscher-etal-2020-zero}.

\noindent
\textbf{Typological Relatedness Features.} Out of all the typological relatedness features, we found Geographical Distance ($d_{geo}(p,t)$) receiving highest SHAP values for all tasks, implying that geographical proximity between the pivot-target pair is an  important factor in determining the zero-shot transferability between them. \citet{lauscher-etal-2020-zero} also observe positive correlations between geographical relatedness and zero-shot performance. The cross-task importance of geographic distance (unlike the other relatedness features)  might be attributed to the 100\% coverage across languages for the geographical vectors in the URIEL database. In contrast, Syntactic and Phonological vectors have missing values for a majority of the languages \cite{littell-etal-2017-uriel}. 

Like \citet{lauscher-etal-2020-zero}, we also see some dependence on syntactic ($s_{syn}(p,t)$) and phonological ($s_{pho}(p,t)$) similarities for XLMR's zero shot performance on XNLI and XQUAD tasks respectively. However, in both cases we found that the tokenizer feature $\text{PCW}(t)$ receives a much higher SHAP value. Interestingly, genetic similarity ($s_{gen}(p,t)$) is not selected for any task, arguably due to the block sparsity in feature selection of Group Lasso. We do see some tasks receiving high SHAP values for $s_{gen}(p,t)$ in single-task lasso (Figure \ref{fig:lasso_xlmr_shap} in Appendix). However, the number of such tasks as well as the SHAP values are on the lower side, implying that genetic similarity might not provide any additional information for zero-shot transfer over and above the geographical, syntactic and phonological similarities.

Similar trends are observed in the case of mBERT as well (Figure \ref{fig:mbert_shap} in appendix), with some minor differences. For instance, instead of $\text{PCW}(t)$, $\text{FERT}(t)$ receives higher SHAP value; $s_{syn}(p,t)$ also receives higher importance, especially for tasks like UDPOS and XNLI, which is consistent with the findings of \citet{lauscher-etal-2020-zero}.



\section{Conclusion and Future Work}

In this paper, we showed that the zero-shot performance prediction problem can be much more effectively and robustly solved by using multi-task learning approaches. We see significant reduction in errors compared to the baselines and single-task models, specifically for the tasks which have test sets available in a very few languages or when trying to predict the performance for low resource languages. Additionally, this approach allows us to robustly identify factors that influence zero-shot performance. Our findings in this context can be summarized as follows. 

\begin{enumerate*}
    \item {\em Subword overlap} between the pivot and target has a strong positive influence on zero-shot transfer, especially for Retrieval tasks.
    \item {\em Quality of the target tokenizer}, defined in terms of how often or how aggressively it splits the target tokens negatively influences zero-shot performance for word-level tasks such as POS tagging and Span extraction. 
    \item {\em Pre-training size} of the target positively influences zero-shot performance in many tasks, including XCOPA, Tatoeba, MLQA and LAReQA.
    \item {\em Geographical proximity} between pivot and target is found to be uniformly important across all the tasks, unlike syntactic and phonological similarities, which are important for only some tasks.
\end{enumerate*}

This last finding is especially interesting. As described earlier, geographical proximity is a more clear, noise-free and complete feature compared to the other relatedness metrics. However, one could also argue that since neighboring languages tend to have high vocabulary and typological feature overlap due to contact processes and shared areal features, geographical distance is an extremely informative feature for zero-shot transfer. Two direct implications of these findings are: (1) for effective use of MMLMs, one should develop resources in at least one pivot language per geographic regions, and (2) one should work towards multilingual tokenizers that are effective for most languages.

There are a number of directions that can be explored in future related to our work. The prediction models can be extended to a multi-pivot and few-shot settings, as described in \citet{srinivasan2021predicting}. Further probing experiments could be designed to understand the role of sub-word overlap on zero-shot transfer of Retrieval tasks. 



\section*{Acknowledgements}
We would like to thank the LITMUS team at Microsoft for their valuable inputs and feedback over the course of this project. 

\bibliography{anthology,custom}
\bibliographystyle{acl_natbib}

\appendix
\section{Appendix}
\label{sec:appendix}

\subsection{Additional Details of Approaches Used}
\label{sec:gpr_maml}
\noindent \textbf{Gaussian Process Regression} (GPR): We start by briefly reviewing Gaussian Processes (GP) in context of the zero-shot performance prediction problem.
For a pivot-target language pair $(p,t)$ and a task $\mathfrak{t}$, the GP prior and the likelihood function can be defined as:
\begin{equation}
    f \sim \mathcal{N}(\mu^{\mathfrak{t}}, K^{\mathfrak{t}}); \,\,\ y|f(x_{p,t}) \sim  \mathcal{N}(y_{p,t}^{\mathfrak{t}}; f(x_{p,t}), \sigma^2_\mathfrak{t})
    \label{eq:gp}
\end{equation}
where  $\mu^{\mathfrak{t}}$ is the mean and $K_{(p,t),(p',t')}^{\mathfrak{t}} = k^{\mathfrak{t}}(x_{p,t}, x_{p',t'})$ is the kernel of the GP defined on the task $\mathfrak{t}$. $\sigma^2_\mathfrak{t}$ denotes the noise variance.

\vspace{1em}

\noindent \textbf{Deep Gaussian Process Regression} (DGPR): We use DGP \cite{pmlr-v51-wilson16} to learn rich features from the observed data. Specifically, the kernel $k^{\mathfrak{t}}(x_{p,t}, x_{p', t'})$ now takes the transformed inputs as
\begin{equation}
k^{\mathfrak{t}}(x_{p,t}, x_{p',t'}) = k^{\mathfrak{t}}(g(x_{p,t}), g(x_{p',t'}))
\end{equation}
where $g(x)$ is a non-linear mapping given by a deep network.
Please refer to \citet{pmlr-v51-wilson16} for a detail account on optimization of DGP.

\vspace{1em}

\noindent \textbf{Multi-Task Deep Gaussian Process Regression (MDGPR)}: We use the multi-task variant of Gaussian Processes proposed in ~\citet{NIPS2007_66368270} where inter-task similarities are learnt solely based on the task identities and the observed data for each task.
Instead of learning task-specific kernels $k^{\mathfrak{t}}(g(x_{p,t}), g(x_{p',t'}))$, we will have a common kernel over the inputs as $k(g(x_{p,t}), g(x_{p',t'}))$ and a positive semi-definite matrix $K_{\text{task}}$ for learning inter-task similarities. Specifically, we define the  multi-task kernel $K_m$ as follows
\begin{multline}
k_m([x_{p,t}, \mathfrak{t}], [x_{p',t'}, \mathfrak{t}']) = \\
k(g(x_{p,t}), g(x_{p',t'})) \ast k_{\text{task}}(\mathfrak{t}, \mathfrak{t}')
\end{multline}

The GP prior will be defined by replacing the task specific kernel $K^{\mathfrak{t}}$ in the equation \ref{eq:gp} with the multi-task kernel $K_m$. We use the  optimization steps similar to DGP and the inference is done by using the standard GP formulae.

Relating MDGPR to equation \ref{eq:mt_obj}, the global parameters $\Theta$ are the parameters of the deep network $g$, and the task specific parameter $\Phi$ is the  positive semi-definite matrix $K_{\text{task}}$.

\noindent
\textbf{Model Agnostic Meta Learning (MAML)}: MAML \cite{finn2017model} is a popular meta learning algorithm that can be used to quickly adapt Deep Neural Networks on new tasks in a few-shot setting. In MAML, the set of initialization parameters for the neural network are explicitly learned such that the network can generalize well on a new task with a small number of gradient steps and training samples.

Relating to equation \ref{eq:mt_obj}, the global parameters $\Theta$ can be considered as the initial set of parameters for the neural network that are learned and shared across all the tasks. Task specific parameters $\Phi$ are adapted from $\Theta$ by taking $K$ gradient steps using the task's performance data.

For evaluating a task $\mathfrak{t}$, we consider rest of the tasks in our dataset as helpers ($\mathfrak{t}' \in \mathfrak{T} - \{\mathfrak{t}\}$) and use them to train the initial set of parameters $\Theta$. The initial parameters are then updated by fine-tuning the network on the training set for $\mathfrak{t}$ using gradient descent.

\subsection{Comparison between mBERT and XLMR Tokenizers}
\label{sec:mmlm_tokenizer}
\begin{figure}[!]
     \begin{subfigure}{.5\textwidth}
         \centering
         \includegraphics[width=.95\linewidth]{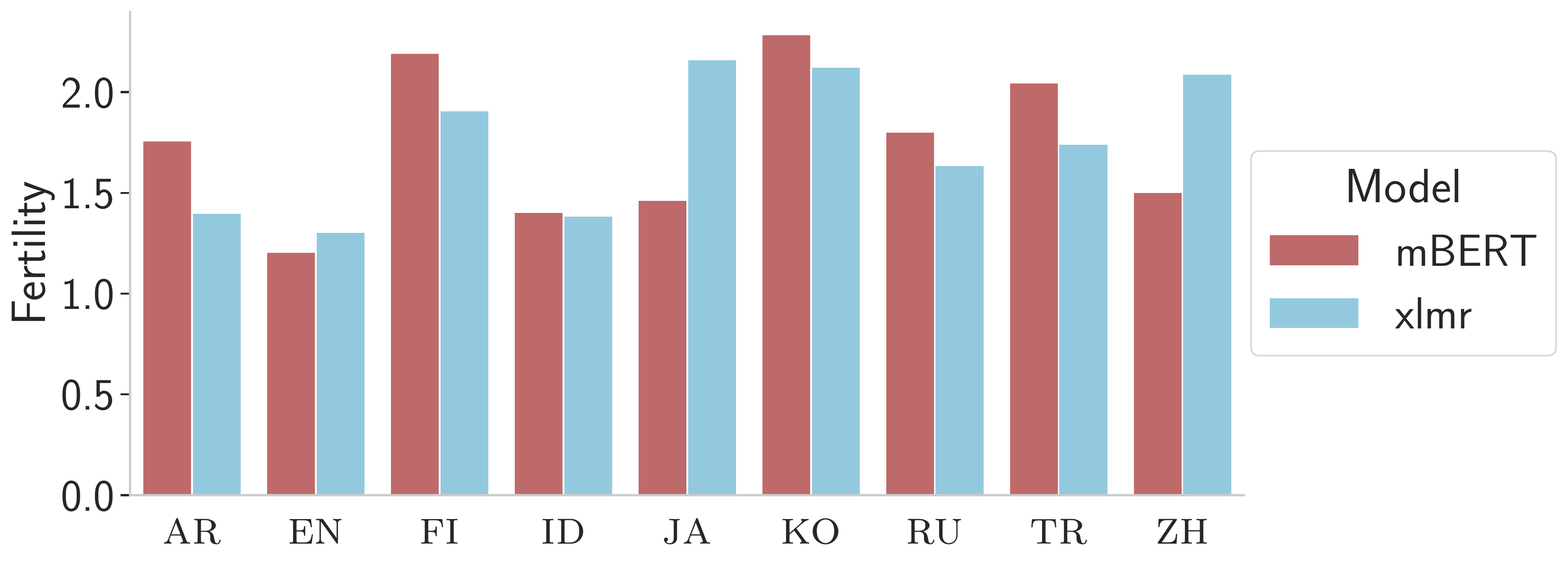}
         \label{fig:fertility}
     \end{subfigure}\\
     \begin{subfigure}{.5\textwidth}
         \centering
         \includegraphics[width=.95\linewidth]{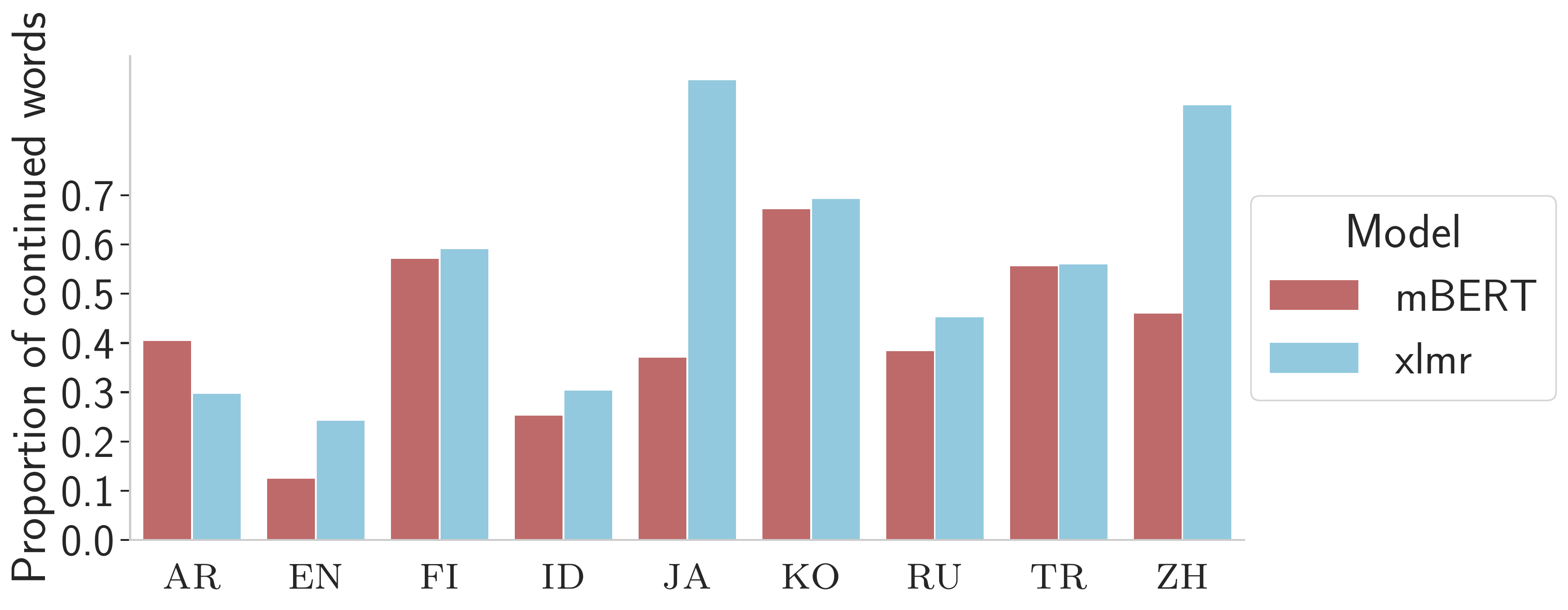}
         \label{fig:pcw}
    \end{subfigure}
 \caption{Comparison of Tokenizer metrics as described by \citet{rust-etal-2021-good} on different languages for MBERT and XLMR. For most languages both model's have similar values of fertility and proportion of continued words, however for Chinese and Japanese the values for XLMR are much higher, which might indicate the subpar quality of XLMR's tokenizer in these languages.}
 \label{fig:tokenizer}
\end{figure}

\begin{figure}[!]
     \begin{subfigure}{.5\textwidth}
         \centering
         \includegraphics[width=.95\linewidth]{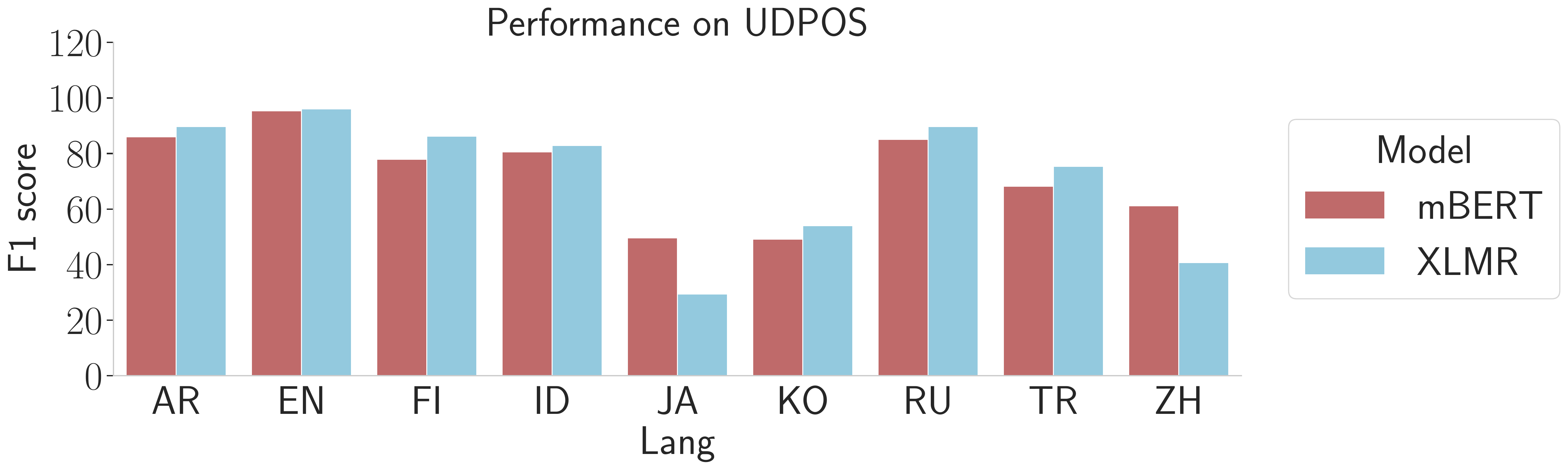}
         \label{fig:udpos}
         \caption{}
     \end{subfigure}\\
     \begin{subfigure}{.5\textwidth}
         \centering
         \includegraphics[width=.95\linewidth]{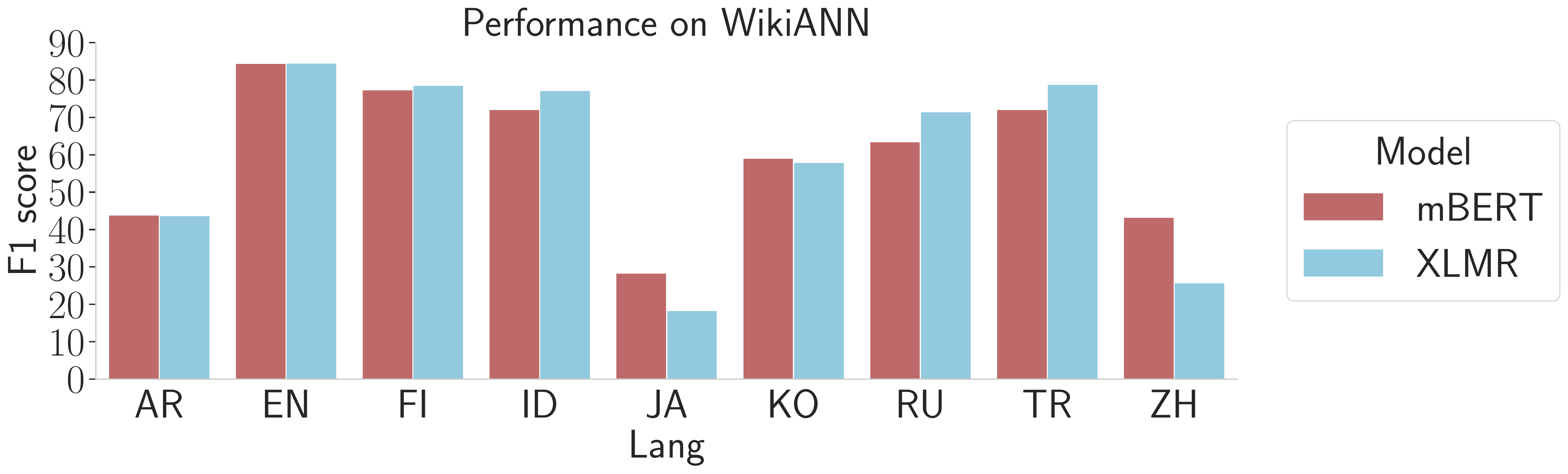}
         \label{fig:ner}
         \caption{}
    \end{subfigure}
 \caption{Zero-shot performance comparison between mBERT and XLMR on (a) UDPOS and (b) WikiANN (NER) tasks, as given in \citet{ruder2021xtreme}}
 \label{fig:mbertVxlmr}
\end{figure}

The $\text{FERT}$ and $\text{PCW}$ metrics as proposed by \citet{rust-etal-2021-good}, have been compared for mBERT and XLMR in figure \ref{fig:tokenizer}. As can be seen, for most languages the metric values are similar across the two tokenizers, however for languages like Chinese and Japanese, there is a dramatic increase in the values for XLMR. Interestingly, when we compare the zero-shot performance between mBERT and XLMR on structure prediction tasks like UDPOS and WikiANN, we see a surprisingly large drop (upto 20\% absolute drop) in the performance for XLMR on these both Chinese and Japanese, whereas usually XLMR outperforms mBERT on these tasks (Refer to figure \ref{fig:mbertVxlmr}). This observation along with the feature importance for the tokenizer features that we observed for Group Lasso (\ref{fig:xlmr_shap}) indicate that tokenizer quality might play some role in the zero-shot transfer capabilities of the multilingual models.

\begin{figure}[!]
    \centering
    \includegraphics[width=0.45\textwidth]{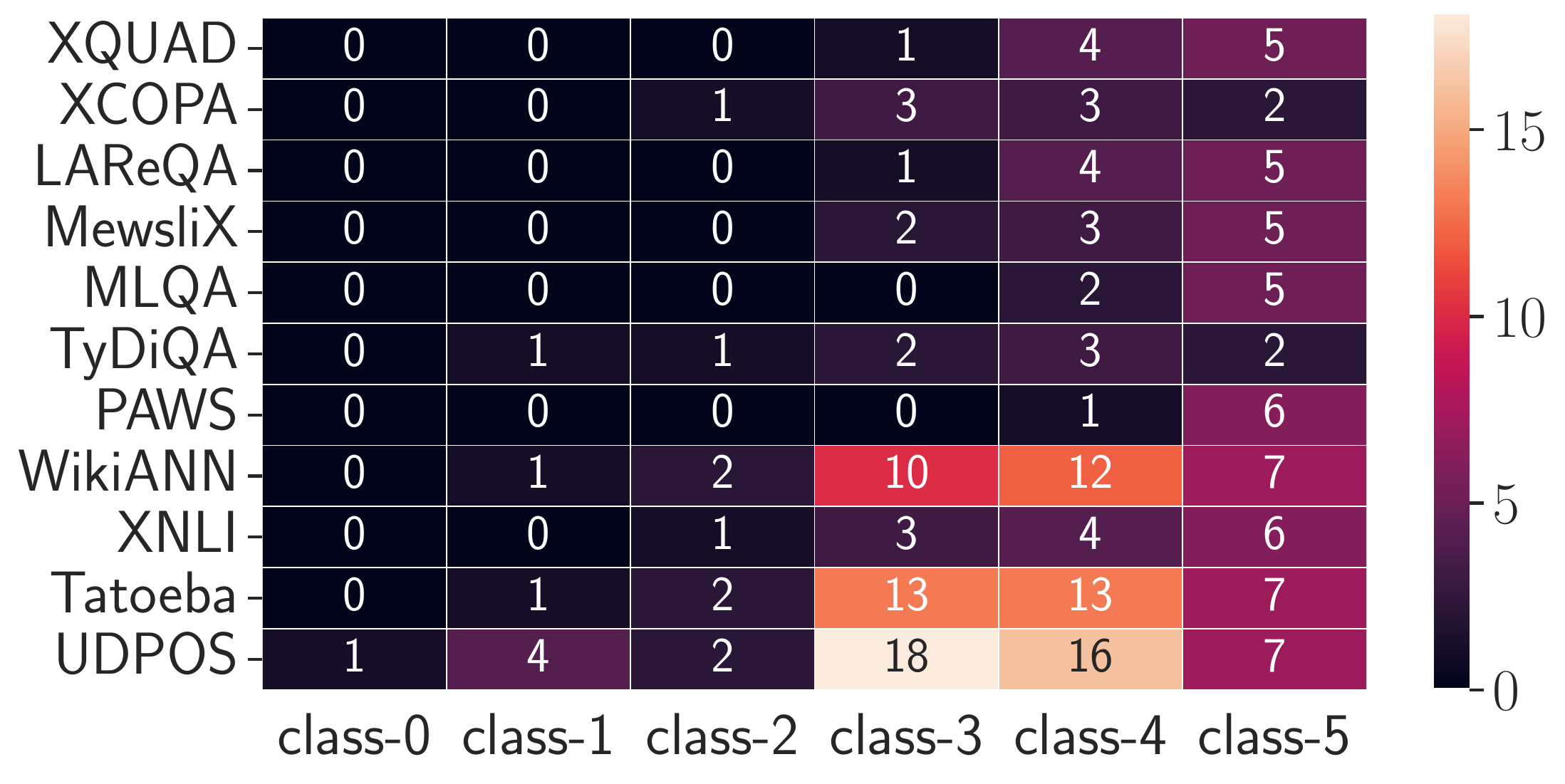}
    \caption{Class wise distribution of languages for different tasks. Languages have been categorized based on the taxonomy provided by \citet{joshi-etal-2020-state}}
    \label{fig:class_dist}
\end{figure}

\begin{figure}[!]
     \includegraphics[width=0.45\textwidth]{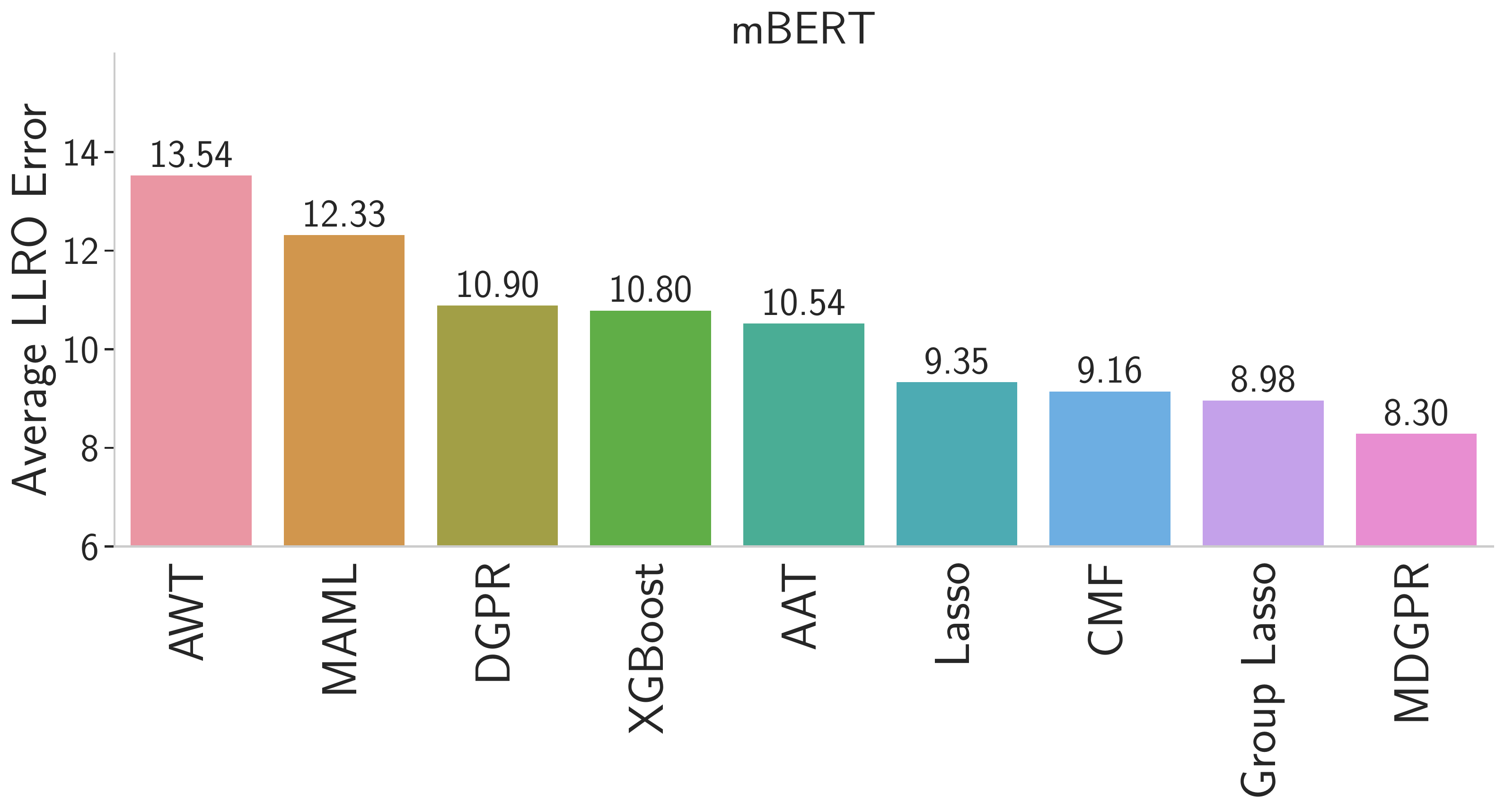}
     \caption{Leave Low Resource Out (LLRO) results for mBERT}
     \label{fig:mbert_lowres}
\end{figure}

\begin{figure}[!]
    \centering
    \includegraphics[width=0.49\textwidth]{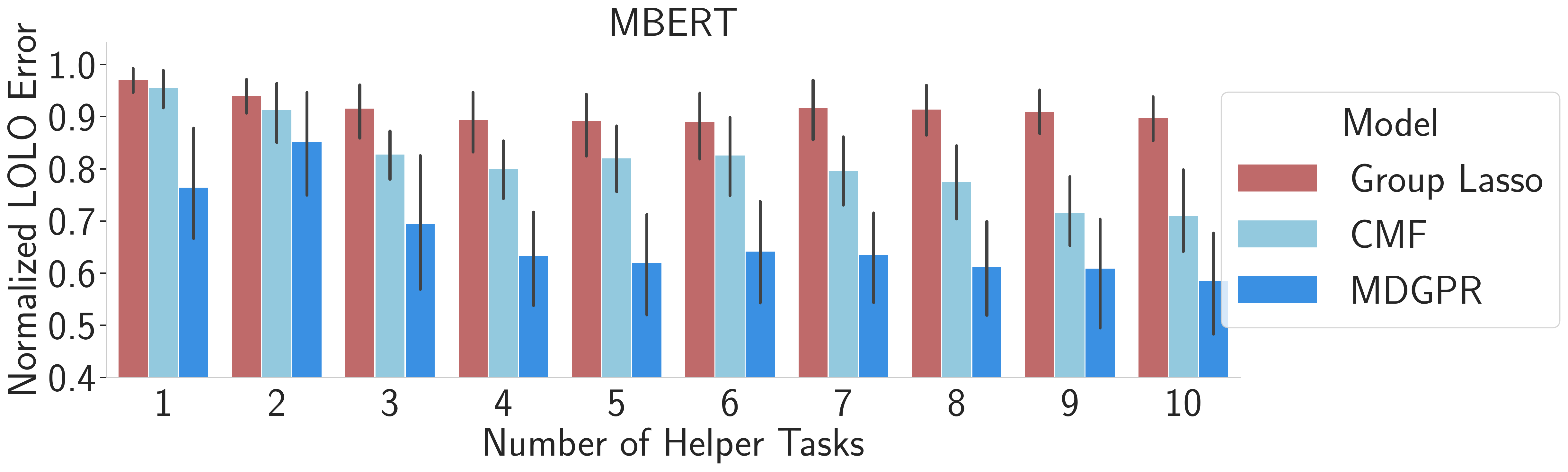}
    \caption{Number of helper tasks vs. LOLO MAE for mBERT. Errors for different model types (Group Lasso, CMF and MDGPR) and tasks are scaled by diving them by the maximum error value.}
    \label{fig:mbert_num_helpers}
\end{figure}

\begin{table*}[]
\scalebox{0.65}{
\begin{tabular}{@{}ccccccccccc@{}}
\toprule
\textbf{Task}    & \textbf{$|\mathcal{T}|$} & \multicolumn{2}{c}{\textbf{Baselines}}                       & \multicolumn{3}{c}{\textbf{Single Task Models}} & \multicolumn{4}{c}{\textbf{Multi Task Models}}                           \\\cmidrule(lr){3-4}\cmidrule(lr){5-7}\cmidrule(lr){8-11}
                 &                                     & \textbf{Average within Task} & \textbf{Average across Tasks} & \textbf{Lasso}         & \textbf{XGBoost} & \textbf{DGPR}   & \textbf{Group Lasso} & \textbf{CMF}  & \textbf{MDGPR} & \textbf{MAML}   \\\midrule
\textbf{MLQA}    & 7                                   & 4.87                                           & 4.59                                            & 6.39                             & 7.47                         & 6.12     & 3.45                                   & 3.18                  & \textbf{2.42}                              & 3.75                            \\
\textbf{PAWS}    & 7                                   & 4.01                                           & 2.96                                           & 3.97                             & 3.01                         &  3.53     & 2.34                                   & 2.75                  & \textbf{1.92}                              & 6.77                            \\
\textbf{XCOPA}   & 8                                   & 3.44                                           & 3.63                                            & 3.54                             & 4.24                      & 3.10         & 3.30                                   & 2.86                  & \textbf{2.59}                              & 5.38                            \\
\textbf{TyDiQA}  & 9                                   & 5.06                                           & 7.08                                            & \textbf{3.42}                    & 6.44                              & 3.94 & 5.09                                   & 4.59                           & 3.92                              & 8.34                            \\
\textbf{XQUAD}   & 10                                  & 6.56                                           & 2.97                                           & \textbf{2.89}                    & 4.69                             & 3.26 & 4.16                                   & 4.37                           & 3.13                              & 4.86                            \\
\textbf{LAReQA}  & 10                                  & 5.57                                           & 2.79                                           & 2.59                             & 4.40                         & 2.64      & 2.22                          & 1.96                           & \textbf{1.75}                              & 8.74                            \\
\textbf{MewsliX} & 10                                  & 19.23                                          & 15.48                                           & 12.15                            & 15.54                     & 17.52         & 10.53                          & \textbf{9.54}                           & 15.99                             & 14.72                           \\
\textbf{XNLI}    & 14                                  & 5.29                                           & 2.94                                           & 3.29                             & \textbf{2.60}               & {2.95}       & 3.18                                   & 3.89                           & 2.98                              & 5.05                            \\
\textbf{WikiANN} & 32                                  & 14.79                                          & 10.54                                           & 9.37                             & 11.13                              & 11.51 & 10.30                                  & 8.91                           & \textbf{8.62}                     & 11.80                           \\
\textbf{Tatoeba} & 35                                  & 14.63                                          & 11.86                                           & \textbf{6.43}                    & 9.57                      & 6.38         & 6.46                          & 7.21                           & \textbf{6.16}                              & 12.13                           \\
\textbf{UDPOS}   & 48                                  & 12.10                                          & 7.43                                           & 7.05                             & {6.37}          & \textbf{6.18}            & 8.94                                   & \textbf{6.58}                           & 6.87                     & 7.97                            \\
\midrule
\textbf{Average} & 19 & 8.69 & 6.57 & 5.55 & 6.86 & 6.10 & 5.45 & \textbf{5.08} & 5.12 & 8.14\\
\textbf{Average ($|\mathcal{T}| \leq 10$)} & 9 & 6.96 & 5.64 & 4.99 & 6.54 & 5.73 & 4.44 & \textbf{4.18} & 4.53 & 7.51\\

\bottomrule
\end{tabular}
}
\caption{\label{tab:results_mbert} Mean Absolute Errors (Scaled by 100 for readability) for different models trained to predict the zero shot performance of mBERT. In the ``Average'' row we average the MAEs across all the tasks and in the ``Average Low'' Res Tasks", we consider the tasks with fewer than 10 target languages and take the average of the MAEs for those tasks.}
\end{table*}

\begin{figure}[!]
    \centering
    \includegraphics[width=0.49\textwidth]{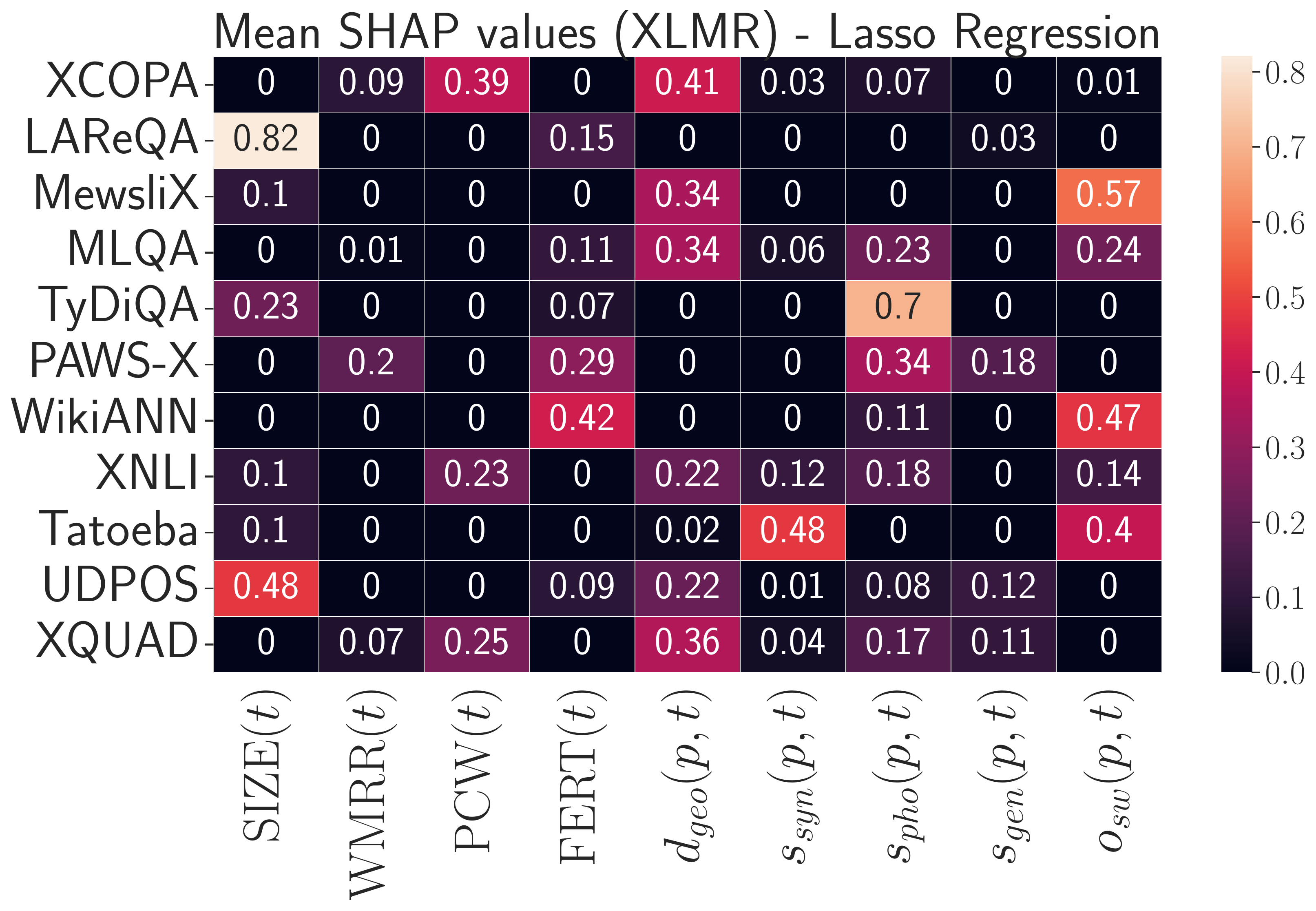}
    \caption{Task-wise mean SHAP values of different features for the Single Task Lasso Regression model trained on XLMR zero-shot performance data.}
    \label{fig:lasso_xlmr_shap}
\end{figure}

\begin{figure}[!]
    \centering
    \includegraphics[width=0.49\textwidth]{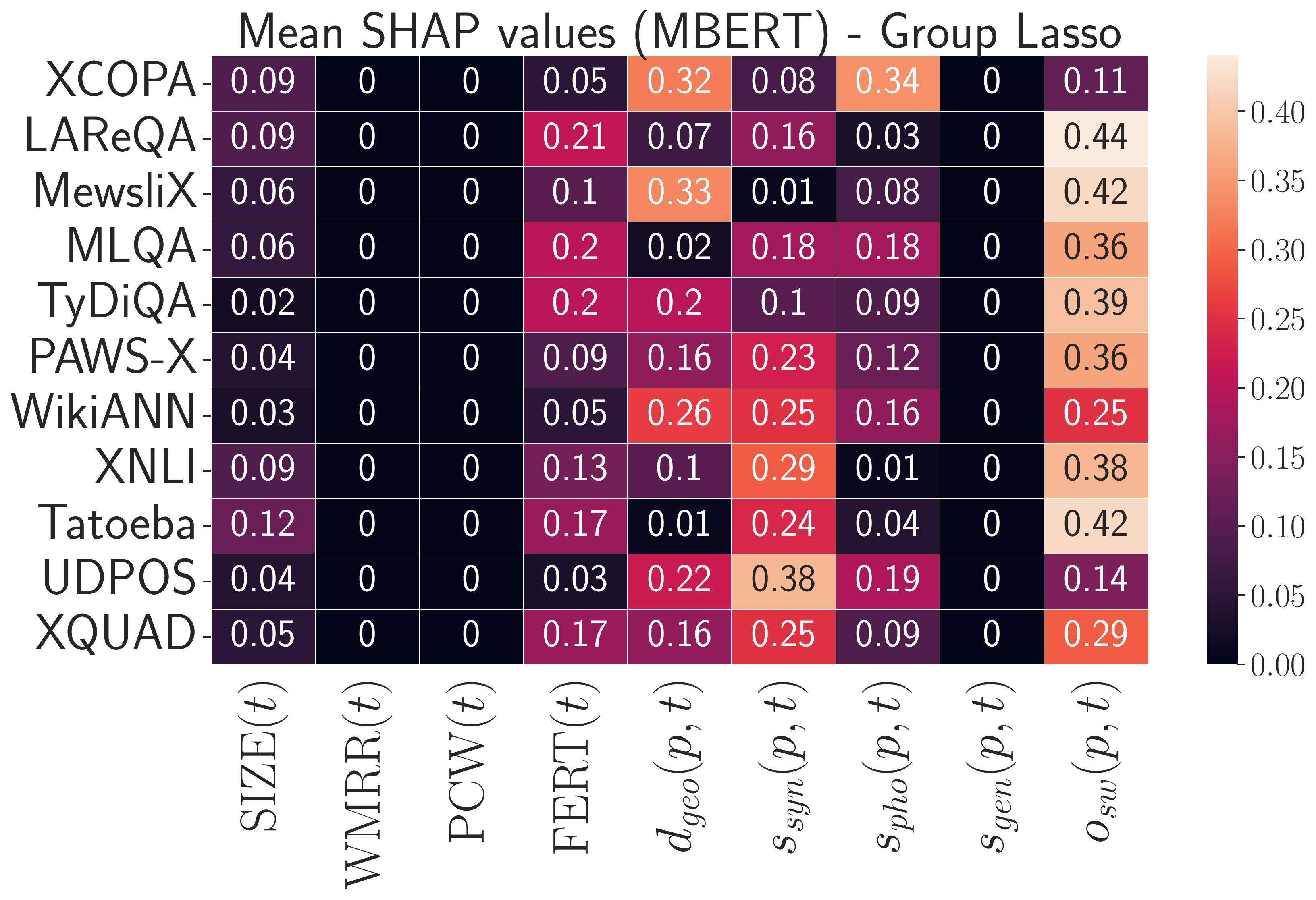}
    \caption{Task-wise mean SHAP values of different features for the Group Lasso model trained on mBERT zero-shot performance data.}
    \label{fig:mbert_shap}
\end{figure}

\begin{figure}[!]
    \centering
    \includegraphics[width=0.49\textwidth]{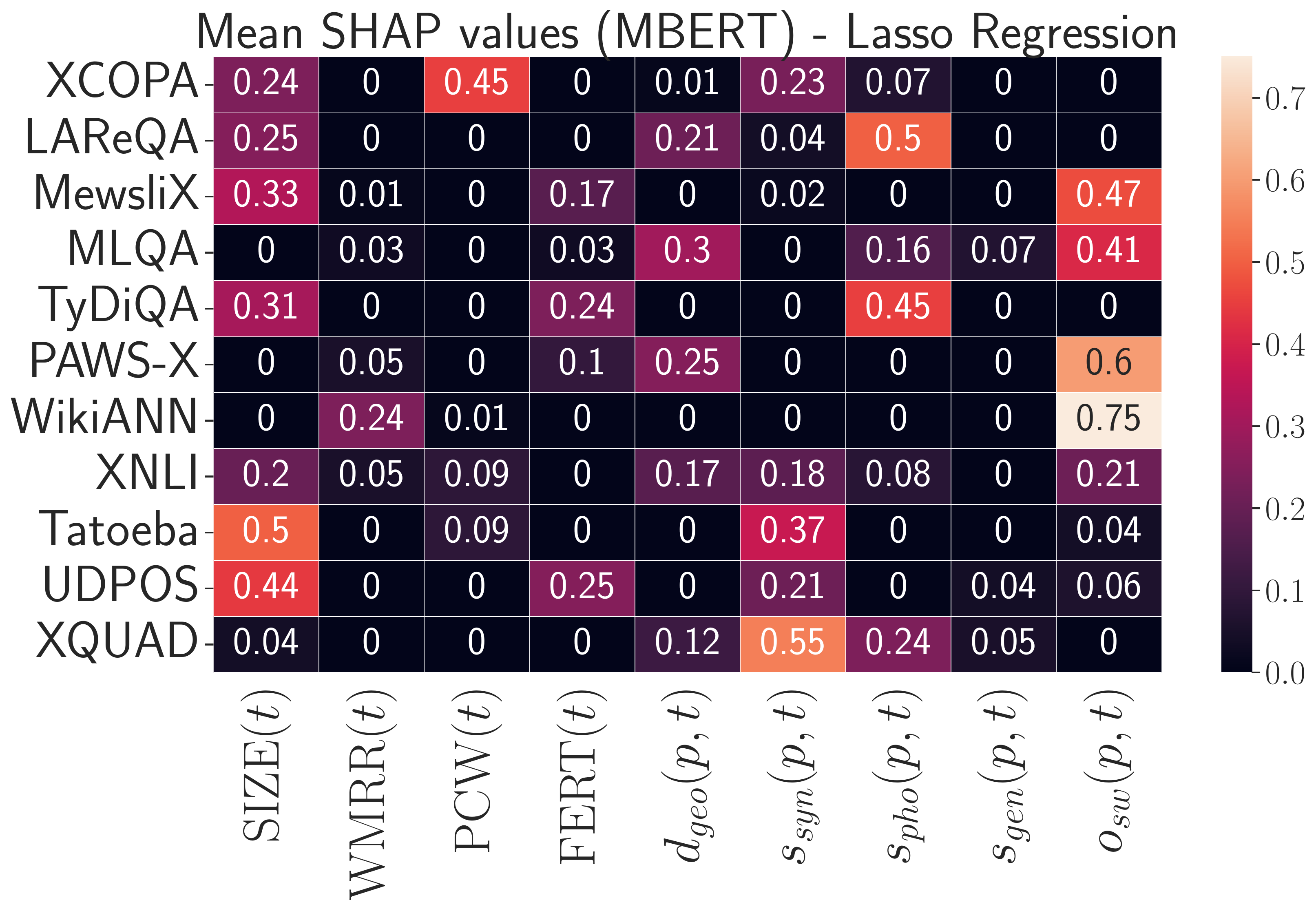}
    \caption{Task-wise mean SHAP values of different features for the Single Task Lasso Regression model trained on mBERT zero-shot performance data.}
    \label{fig:lasso_mbert_shap}
\end{figure}

\end{document}